\documentclass[letterpaper]{article} 
\usepackage{aaai2027}  

\nocopyright

\usepackage[hyphens]{url}  
\usepackage{graphicx} 
\usepackage{natbib}  
\usepackage{caption} 
\usepackage{algorithm}
\usepackage{algorithmic}

\usepackage{amsmath}
\usepackage{amsfonts}
\usepackage{multirow}

\usepackage{newfloat}
\usepackage{listings}
\DeclareCaptionStyle{ruled}{labelfont=normalfont,labelsep=colon,strut=off} 
\floatstyle{ruled}
\newfloat{listing}{tb}{lst}{}
\floatname{listing}{Listing}

\usepackage{booktabs}

\title{DS-VLA: A Dendritic-inspired Vision-Language-Action Model \\for Robust Action Control}

\author{
    Yaxing Lyu\textsuperscript{\rm 1,\rm 3},
    Jingyi Li\textsuperscript{\rm 1,\rm 4},
    Mingkun Xu\textsuperscript{\rm 1}\corresponding,
    Yujie Wu\textsuperscript{\rm 2}\corresponding
}

\affiliations{
    \textsuperscript{\rm 1}Guangdong Institute of Intelligence Science and Technology\\
    \textsuperscript{\rm 2}Hong Kong Polytechnic University\\
    \textsuperscript{\rm 3}The University of Hong Kong\\
    \textsuperscript{\rm 4}University of Illinois Urbana-Champaign\\
    yu-jie.wu@polyu.edu.hk, lyuyaxing@connect.hku.hk
}

\begin{document}

\maketitle

\begin{abstract}
Vision–language–action (VLA) models have achieved strong performance in language-conditioned manipulation, yet success under nominal evaluation does not necessarily translate into robust closed-loop behavior when executed actions are transiently corrupted. We introduce DS-VLA, a dendritic-inspired action architecture that incorporates dendritic spiking dynamics into VLA control to address this limitation. Specifically, to enable modularized feature processing and temporal information integration, DS-VLA equips action neurons with multiple sparsely connected dendritic branches, each featuring heterogeneous, learned decay factors. Furthermore, to suppress unreliable state updates while preserving task-relevant historical information, we introduce a neuron-wise inhibitory gate that adaptively regulates the admission of new multimodal evidence into dendritic states prior to somatic dynamics. We evaluate DS-VLA on all four LIBERO suites under both nominal rollouts and a unified closed-loop action-perturbation protocol. DS-VLA achieves a 91.6\% average nominal success rate and an 87.35\% average perturbed success rate, retaining 95.4\% of its nominal performance. Under the same reported perturbation setting, OpenVLA-OFT, FAST, $\pi_0$, and GR00T achieve 39.45\%, 23.90\%, 28.55\%, and 30.75\%, respectively. A controlled ablation isolates the contribution of neuron-wise shared inhibition, while analyses of neural dynamics and post-perturbation trajectories associate robust performance with selective evidence suppression and effective behavioral recovery. Together, these results demonstrate that integrating brain-inspired computational mechanisms offers a promising architectural prior for robust embodied intelligence beyond merely scaling vision–language backbones or generative action decoders.
\end{abstract}

\begin{links}
    \link{Code}{https://github.com/ASTAR123/DS-VLA}
\end{links}

\section{Introduction}
\label{sec:introduction}

\begin{figure}[t]
    \centering
    \includegraphics[width=0.98\columnwidth]{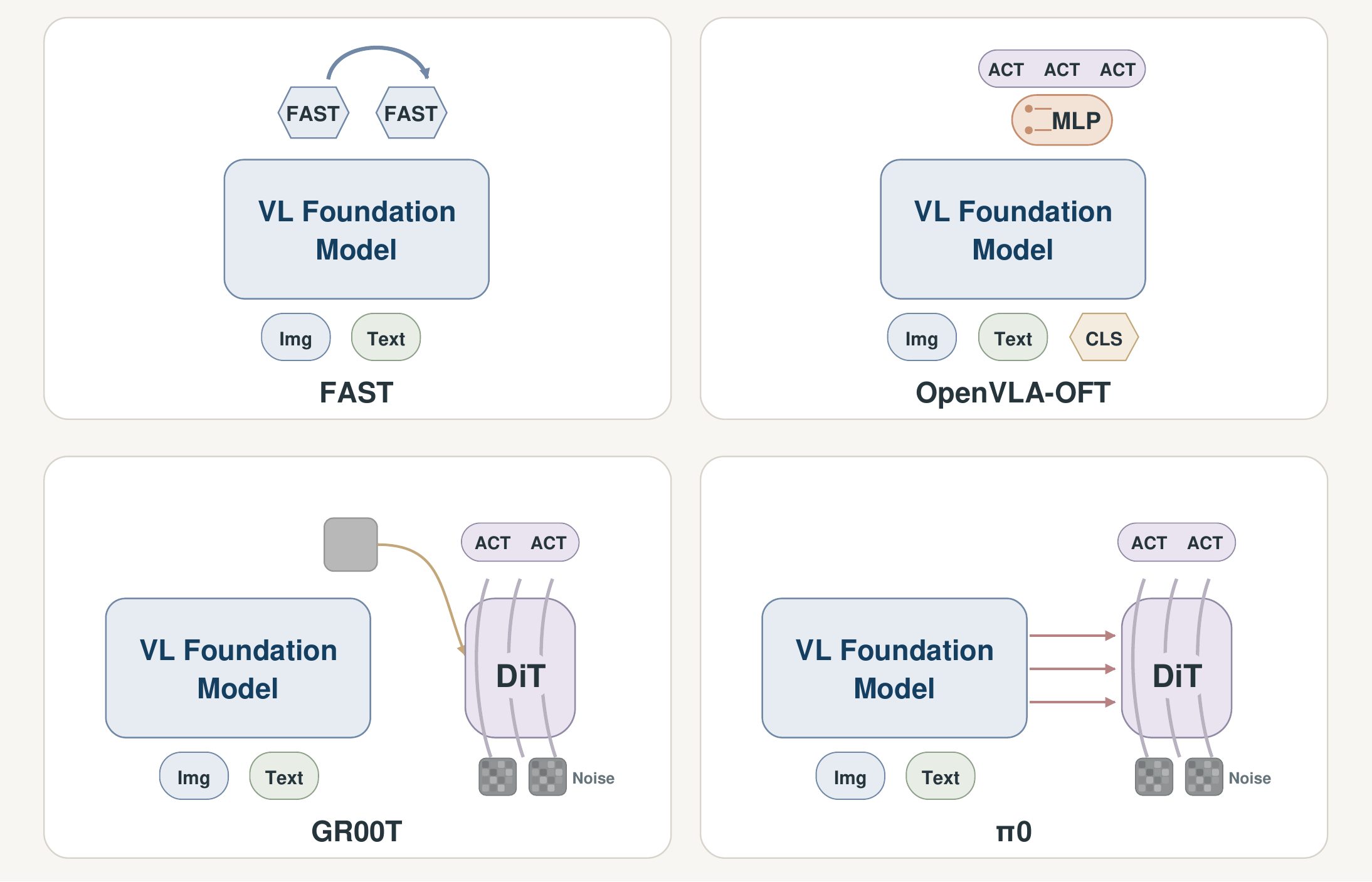}
    \caption{Representative VLA action-modeling paradigms under a shared multimodal interface. FAST autoregressively decodes frequency-space action tokens, OpenVLA-OFT predicts continuous action chunks in parallel, $\pi_0$ generates actions through flow matching, and GR00T couples semantic reasoning with real-time motor generation through a dual-system architecture.}
    \label{fig:vla_action_heads}
\end{figure}

Vision--language--action (VLA) models connect multimodal foundation models with embodied control by mapping visual observations and language instructions to robot actions~\cite{zitkovich2023rt,kim2024openvla}. In recent years, progress in this field has been driven primarily by larger-scale pretraining data, stronger multimodal backbones, and advances in action modeling, exemplified by the autoregressive frequency-space action tokenization of FAST~\cite{pertsch2025fast}, the parallel continuous action-chunk regression of OpenVLA-OFT~\cite{kim2025fine}, the flow-matching action expert of $\pi_0$~\cite{black2024pi_0}, and the dual-system architecture of GR00T with a diffusion-transformer action module for real-time motor generation~\cite{bjorck2025gr00t}.  Fig.~\ref{fig:vla_action_heads} contrasts these four representative designs under a common vision--language foundation-model interface~\cite{community2026starvla}. These advances have  improved manipulation performance under standard evaluation, but nominal task success does not guarantee closed-loop robustness. Action noise, actuator errors, unexpected contact, or incorrect gripper states can alter both robot and object states, causing subsequent observations to deviate from the intended trajectory. A robust policy must therefore preserve task intent, limit disruptive updates to its internal state, and adapt its actions to the changed environment.

\begin{figure}[t]
    \centering
    \includegraphics[width=0.98\columnwidth]{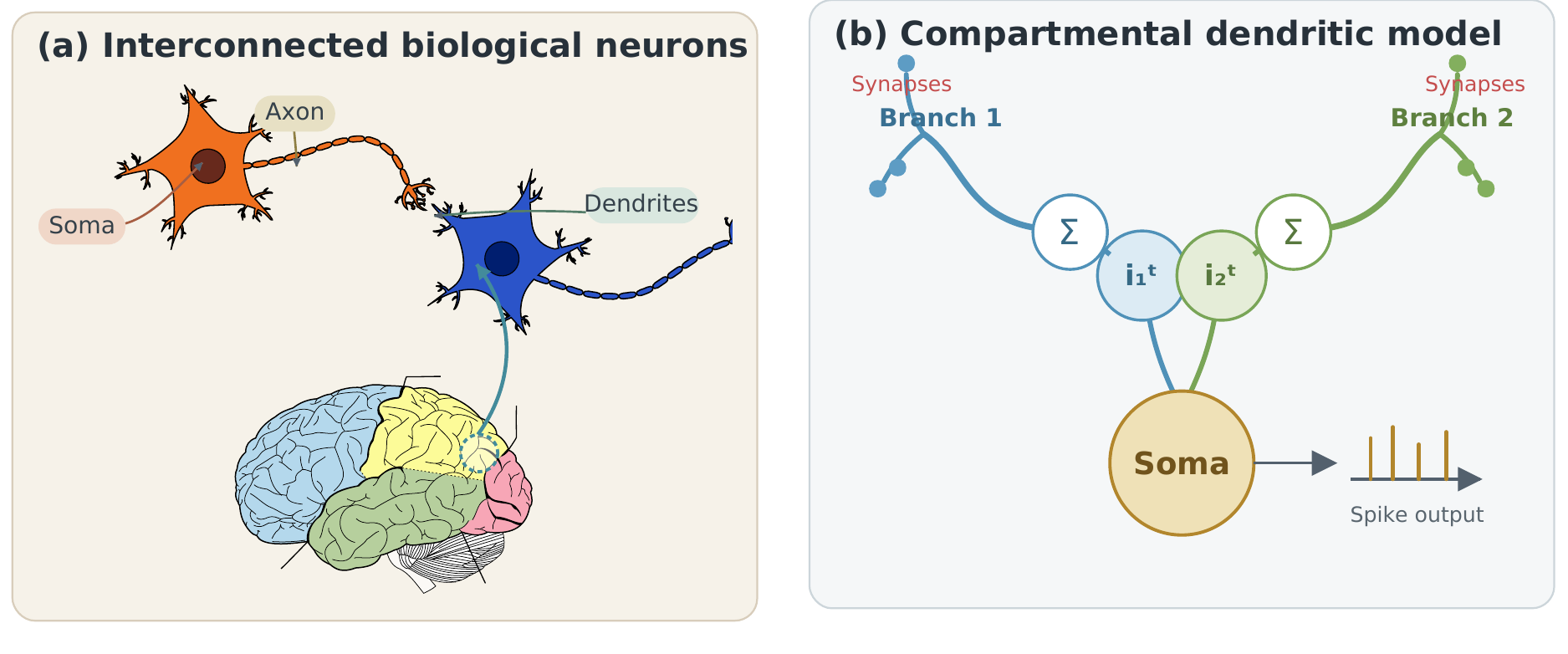}
    \caption{From interconnected biological neurons to computational dendritic modeling. Top: two neurons connected through an axon--dendrite synapse. Bottom: synaptic inputs are integrated by separate dendritic branches before converging at the somatic unit.}
    \label{fig:dendritic_neuron_model}
\end{figure}

Biological neural systems offer useful principles for constructing such action dynamics. Conventional artificial neural networks typically abstract a neuron as a point-like computational unit. In contrast, biological pyramidal neurons integrate inputs from different pathways locally within dendritic compartments before combining branch responses at the soma~\cite{hausser2000diversity, gasparini2006state}. Dendritic branches exhibit diverse connectivity and temporal response properties, allowing a single neuron to organize information across input subspaces and timescales. In spiking neural networks, explicitly modeling temporal heterogeneity across dendritic compartments has likewise been shown to support multi-timescale feature integration and improve robustness and generalization on temporal tasks~\cite{zheng2024temporal}. As illustrated in Fig.~\ref{fig:dendritic_neuron_model}, our abstraction maps this biological compartmentalization to sparse branch-specific receptive fields, distinct dendritic states, and somatic integration. Input partitioning can restrict the spread of local disturbances, dendritic state can attenuate transient fluctuations, and heterogeneous branch dynamics can jointly represent rapid changes and persistent task context. Dendritic compartments are therefore not merely a structural extension of point neurons; their local memory, temporal filtering, and branch-wise integration provide a natural dynamical prior for robust action modeling.

Cortical inhibitory interneurons further regulate dendritic processing in an activity-dependent and compartment-specific manner~\cite{palmer2012inhibitory}. Rather than uniformly silencing an entire network, inhibitory circuits can selectively modulate particular neurons or dendritic regions. Martinotti cells, for example, provide dendrite-targeting inhibition to excitatory neurons, whereas neurogliaform cells can induce longer-lasting local GABAergic modulation through volume transmission~\cite{walker2016parvalbumin,olah2009regulation}. Physiological and computational evidence shows that dendritic inhibition can exert a disproportionately strong influence on dendritic excitability, suppress NMDA-receptor-dependent nonlinearities, and gate the electrogenesis underlying burst firing~\cite{goetz2015dendritic,lovett2012regulation}. These findings motivate controlling the current admitted into the dendritic state in DS-VLA. Its proposed roles in selecting action-relevant evidence and limiting perturbation-induced state updates are computational hypotheses evaluated empirically in this work.

Motivated by these principles, we propose \textbf{DS-VLA}, a dendritic-inspired Vision--Language--Action model for robust action control that introduces dendritic compartmentalization, spiking dynamics, and inhibitory modulation into VLA action generation. Multiple dendritic branches partition and integrate incoming features with distinct temporal responses, while somatic spiking dynamics organize information across time for continuous action prediction. DS-VLA further employs a neuron-wise shared inhibitory gate: an independent, input-dependent inhibitory coefficient is generated for each neuron and shared by its two dendritic branches. It reduces interference from task-irrelevant or phase-inconsistent evidence during nominal execution and limits anomalous state updates under perturbations, allowing the action module to protect accumulated task information while remaining responsive to meaningful environmental changes. DS-VLA focuses on how information enters, updates, and persists in the action state, providing a brain-inspired framework for studying the relationship between neural dynamics and embodied control.

Our main contributions are summarized as follows:
\begin{itemize}
    \item We propose DS-VLA, a dendritic spiking action architecture for continuous robot control that effectively integrates neural information processing mechanisms into VLA backbones, providing a novel solution for improving the closed-loop stability of VLA policies.

    \item We formalize and integrate two neural-inspired mechanisms within the VLA action generation process: branch-specific dendritic integration for multi-timescale feature processing, and neuron-wise shared inhibitory gating for selectively regulating state updates. The proposed unified design supports both effective information processing during nominal tasks and stable state retention under environmental perturbations. 

    \item We establish a systematic evaluation framework covering standard tasks, multiple perturbation levels, and controlled architectural ablation to examine task performance, perturbation resilience, and the role of inhibitory gating. Crucially, DS-VLA retains over 95\% of its nominal performance under standardized action perturbations, significantly outperforming the evaluated baselines.
\end{itemize}

\section{Related Work}
\label{sec:related_work}

\subsection{Vision--Language--Action Models}
Large language models (LLMs) established scalable sequence modeling and transferable linguistic reasoning from large text corpora~\cite{llama2023llama}. Vision--language models (VLMs) extended this paradigm by aligning visual and textual representations~\cite{radford2021learning} and connecting visual encoders with instruction-tuned language models for multimodal reasoning~\cite{liu2023visual}. Recent VLM research further moves from passive observation toward process-level reasoning and active critique in robotic manipulation~\cite{liu2026passive}. VLA models further ground these representations in control: RT-2 casts robot actions as tokens within a VLM~\cite{zitkovich2023rt}, while OpenVLA provides an open pretrained VLA for manipulation~\cite{kim2024openvla}. Subsequent work has diversified action generation through frequency-space autoregressive tokenization in FAST~\cite{pertsch2025fast}, parallel continuous action-chunk regression in OpenVLA-OFT~\cite{kim2025fine}, flow matching in $\pi_0$~\cite{black2024pi_0}, and dual-system inference in GR00T~\cite{bjorck2025gr00t}. StarVLA organizes such components into a modular development framework~\cite{community2026starvla}. Complementarily, NeuroVLA introduces a system-level neuromorphic hierarchy that associates high-level planning, adaptive stabilization, and fast spiking execution with cortex-, cerebellum-, and spinal-cord-like modules~\cite{guo2026brain}.

\subsection{Closed-Loop Robustness in VLA Control}

Robot policies operate in a closed loop, where an erroneous action can alter subsequent robot, object, and sensory states. Conventional robustness techniques improve generalization by randomizing visual appearance or physical dynamics during training~\cite{tobin2017domain,peng2018sim}, while rapid motor adaptation infers environment-dependent control variables online~\cite{kumar2021rma}. Generative controllers such as Diffusion Policy model multimodal action distributions and improve visuomotor behavior on contact-rich tasks~\cite{chi2025diffusion}, but expressive action generation alone does not guarantee recovery after execution deviates from the nominal trajectory. STRONG-VLA uses a perturbation curriculum followed by clean-data refinement to improve robustness to visual and linguistic corruptions~\cite{xie2026strong}. RobustVLA evaluates perturbations in both VLA inputs and actions, and improves action robustness through optimization against worst-case action noise~\cite{zhang2025robustvla}. Because an action disturbance changes the physical state encountered at the next control step, robustness also depends on how the policy adapts its subsequent actions to new observations. We study this closed-loop recovery process through action-head dynamics and task success over the resulting trajectory.

\subsection{Spiking Neural Networks, Dendritic Computation, and Inhibitory Modulation}

Spiking neural networks (SNNs) maintain state through membrane dynamics and communicate with discrete events; surrogate-gradient methods make these temporal models trainable with gradient-based optimization~\cite{neftci2019surrogate}. Most artificial spiking neurons nevertheless remain point-neuron abstractions. Biological dendrites perform branch-local integration with nonlinear and state-dependent responses~\cite{hausser2000diversity,gasparini2006state}, motivating computational models that distribute inputs and temporal constants across dendritic compartments. Temporal dendritic heterogeneity has been shown to improve multi-timescale learning and robustness in SNNs~\cite{zheng2024temporal}. Dendritic processing is further shaped by spatially and temporally specific inhibition in vivo~\cite{palmer2012inhibitory}; even sparse dendritic inhibitory input can exert a strong effect on local excitability~\cite{goetz2015dendritic}. Dendrite-targeting interneurons regulate branch responses, while neurogliaform-mediated GABAergic volume transmission provides longer-lasting local modulation~\cite{walker2016parvalbumin,olah2009regulation}. These findings support input-dependent, branch-related regulation as a computational principle. Prior work has largely studied these mechanisms in neuroscience, whereas their computational modeling and practical use in embodied intelligence have yet to be systematically investigated.

\section{Method}
\label{sec:method}

\begin{figure*}[t]
    \centering
    \includegraphics[width=0.85\textwidth]{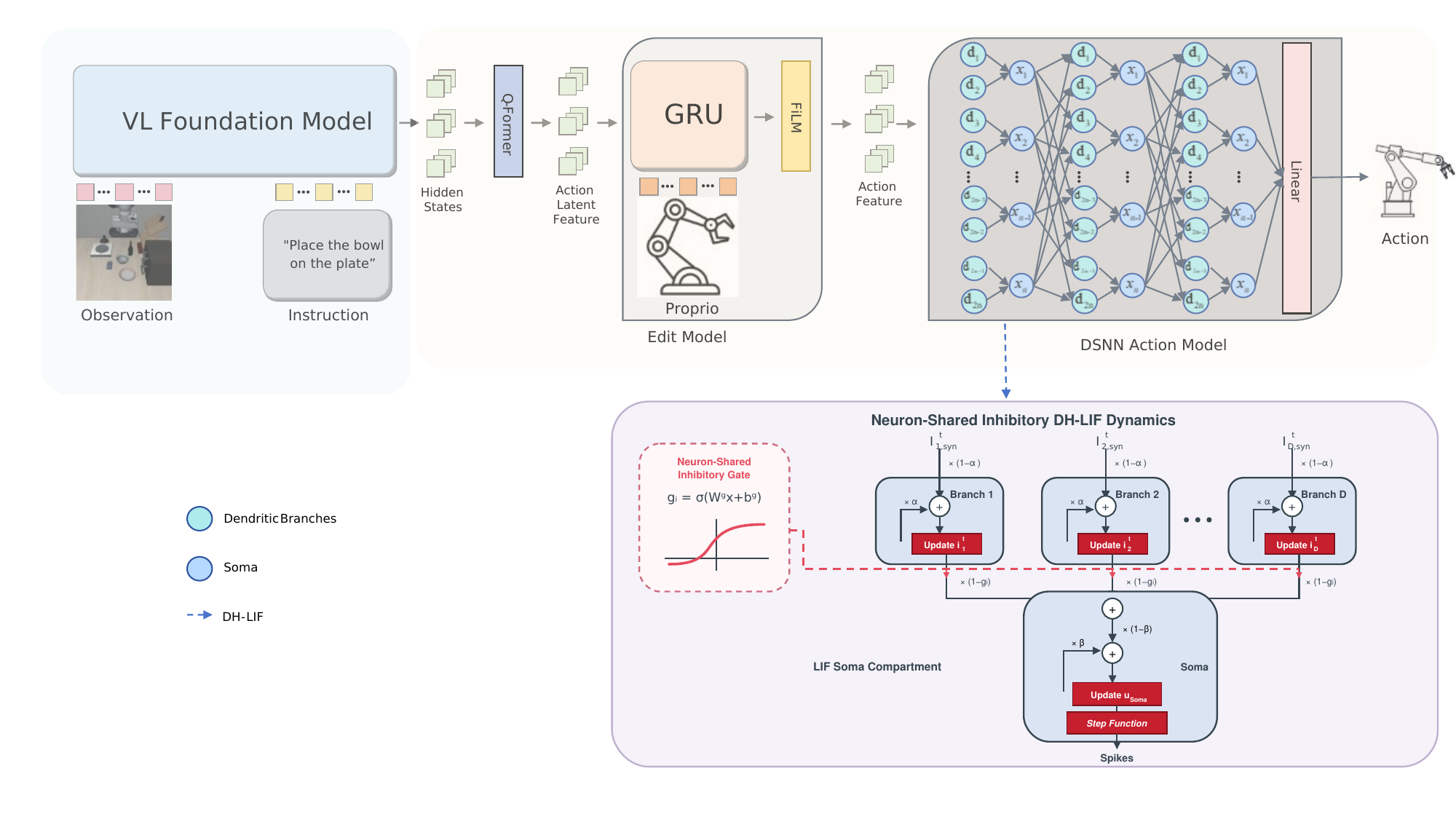}
    \caption{Overview of DS-VLA. Visual observations and language instructions are encoded by the vision--language backbone, and action queries are fused with robot-state features through GRU-based encoding and FiLM modulation. The resulting action features are processed by the dendritic spiking action model and decoded into continuous robot actions. The lower panel details the neuron-shared inhibitory DH-LIF dynamics: each somatic neuron assigns one input-dependent inhibitory coefficient to its two dendritic branches, whose temporally integrated currents subsequently converge at the LIF soma to generate spikes.}
    \label{fig:ds_vla_overview}
\end{figure*}

Given visual observations $I$, a language instruction $c$, and a robot-state history $S=\{s_h\}_{h=1}^{H}$, DS-VLA predicts a continuous action sequence $\widehat{A}=\{\widehat{a}_t\}_{t=1}^{T}$ with $\widehat{a}_t\in{\rm R}^{7}$. As shown in Figure~\ref{fig:ds_vla_overview}, the VLM and action queries extract multimodal features, while a  Gated Recurrent Unit (GRU)~\cite{chung2014empirical} and Gated Feature-wise Linear Modulation (FiLM)~\cite{perez2018film} inject robot-state information before the dendritic spiking action head. Within each DH-LIF layer, sparse dendritic branches retain independent temporal states, and one input-dependent inhibitory coefficient is shared by the two branches of each somatic neuron before their currents converge at the soma. The overall computation is summarized as
\begin{equation}
\begin{array}{c}
Z={\cal Q}({\cal V}(I,c)),\\
X^{(r)}={\cal M}(Z,S^{(r)}),\\
\widehat{A}^{(r)}={\cal F}_{\rm DS}(X^{(r)}).
\end{array}
\label{eq:overview}
\end{equation}
where ${\cal V}$ is the VLM encoder, ${\cal Q}$ extracts action queries, ${\cal M}$ performs state-conditioned modulation, and ${\cal F}_{\rm DS}$ is the dendritic spiking action head. The superscript $r\in\{0,1\}$ indexes the two prediction rounds, and $X^{(r)}$ denotes the state-conditioned action feature at round $r$. The first prediction updates the robot-state tensor through ${\cal U}$, $S^{(1)}={\cal U}(\widehat{A}^{(0)},S^{(0)})$, and the second produces the final sequence $\widehat{A}=\widehat{A}^{(1)}$.

\subsection{Multimodal and State-Conditioned Features}

Selected VLM hidden states $H_l$ are projected to the action dimension, $K_l=H_lW_p+b_p$. Eight learnable queries then extract action-relevant information through successive cross-attention blocks:
\begin{equation}
\begin{array}{c}
Q'_l=Q_{l-1}+{\rm MHA}({\rm LN}(Q_{l-1}),K_l,K_l),\\
Q_l=Q'_l+{\rm MLP}({\rm LN}(Q'_l)).
\end{array}
\label{eq:query_extraction}
\end{equation}
Here, $l$ indexes the selected VLM layers and their corresponding cross-attention blocks; $Q_0\in{\rm R}^{B\times8\times768}$ denotes the eight learned action queries, and $Q_l$ is their updated representation at block $l$. ${\rm MHA}$, ${\rm LN}$, and ${\rm MLP}$ denote multi-head attention, layer normalization, and the feed-forward block, respectively. The projected VLM features $K_l$ serve as both keys and values, and the final query representation is $Z=Q_L$.

The final representation $Z\in{\rm R}^{B\times8\times768}$ is modulated by the robot state. A two-layer unidirectional GRU encodes the state history into its final hidden state $h_S$, while the eight query features are mean-pooled as $\bar z=\frac{1}{8}\sum_{q=1}^{8}Z_q$. Their projected features are
\begin{equation}
p_Z=\phi(W_Z{\rm LN}(\bar z)+b_Z),\quad
p_S=\phi(W_S{\rm LN}(h_S)+b_S),
\label{eq:state_features}
\end{equation}
where $B$ is the batch size, $\phi$ is ReLU, and all $W$ and $b$ terms are learned affine-projection parameters. A state gate forms $p'_Z=p_Z\odot\sigma(W_g p_S+b_g)$, where $\sigma$ is the sigmoid function and $\odot$ denotes element-wise multiplication. Two MLPs then predict the FiLM scale $\gamma$ and shift $\beta$ from the concatenated feature $[p'_Z;p_S]$. The modulated query is
\begin{equation}
X_q=Z_q\odot(1+\gamma)+\beta.
\label{eq:film}
\end{equation}
This feature-fusion gate is distinct from the neuron-wise inhibitory gate defined below.

\subsection{Dendritic Spiking Action Head}

\subsubsection{Sparse Dendritic Integration and Inhibitory Gating}

The action head contains three dendritic spiking layers. Each somatic neuron has two dendritic branches that receive complementary input subsets through a fixed binary mask. For neuron $i$, branch $b$, layer $l$, and internal step $t$, the branch current is
\begin{equation}
I_{i,b,t}^{l}
=\sum_j M_{i,b,j}^{l}W_{i,b,j}^{l}x_{j,t}^{l}+c_{i,b}^{l},
\label{eq:dendritic_current}
\end{equation}
where $j$ indexes input features, $x_{j,t}^{l}$ is the input to layer $l$, $M_{i,b,j}^{l}\in\{0,1\}$ is the fixed connectivity mask, and $W_{i,b,j}^{l}$ and $c_{i,b}^{l}$ are the learned weight and bias.

Each neuron independently computes an input-dependent inhibitory coefficient from the complete layer input:
\begin{equation}
\begin{array}{c}
\displaystyle
g_{i,t}^{l}=\sigma\!\left(\sum_jW_{i,j}^{g,l}x_{j,t}^{l}+b_i^{g,l}\right),\\
\widetilde I_{i,b,t}^{l}=(1-g_{i,t}^{l})I_{i,b,t}^{l}.
\end{array}
\label{eq:inhibitory_gate}
\end{equation}
Here, $W^{g,l}$ and $b^{g,l}$ are the learned gate parameters. The same $g_{i,t}^{l}\in(0,1)$ modulates both branches of neuron $i$, but different neurons retain independent coefficients. Thus, the gate controls how strongly each neuron admits new evidence while preserving the relative organization of its two branch currents. Larger $g$ denotes stronger inhibition.

\subsubsection{Dendritic and Somatic Dynamics}

Each branch maintains a state with a learnable decay $\beta_{i,b}^{d,l}=\sigma(\tau_{i,b}^{l})$:
\begin{equation}
d_{i,b,t}^{l}
=\beta_{i,b}^{d,l}d_{i,b,t-1}^{l}
+(1-\beta_{i,b}^{d,l})\widetilde I_{i,b,t}^{l}.
\label{eq:dendritic_dynamics}
\end{equation}
Branch-specific decay factors provide heterogeneous temporal responses. Importantly, inhibition affects only the incoming-current term and therefore limits anomalous updates without erasing stored dendritic state.

The soma sums its two branch states and applies a bounded membrane skip. Let $\lambda_i^l$ be the learnable LIF somatic decay and $\alpha^l=0.2\sigma(\rho^l)$ the layer-wise skip coefficient. The pre-spike membrane potential is
\begin{equation}
\widetilde u_{i,t}^{l}
=\lambda_i^l u_{i,t-1}^{l}
+\sum_b d_{i,b,t}^{l}
+\alpha^l u_{i,t-1}^{l}.
\label{eq:somatic_integration}
\end{equation}
where $u_{i,t-1}^{l}$ is the post-reset membrane state from the preceding internal step, $\sum_b d_{i,b,t}^{l}$ aggregates the two dendritic states, and $\rho^l$ is a learned skip logit. The decay $\lambda_i^l$ is initialized in $[0.95,0.99]$ and learned independently for each somatic neuron.
Spike generation and subtractive reset follow
\begin{equation}
\begin{array}{c}
z_{i,t}^{l}={\cal H}(\widetilde u_{i,t}^{l}-\vartheta_i^l),\\
u_{i,t}^{l}=\widetilde u_{i,t}^{l}-z_{i,t}^{l}\vartheta_i^l.
\end{array}
\label{eq:lif_update}
\end{equation}
where ${\cal H}$ is the Heaviside step function, $\vartheta_i^l$ is the learnable firing threshold, and $z_{i,t}^l\in\{0,1\}$ is the output spike. A fast-sigmoid surrogate derivative is used during backpropagation.

\subsection{Continuous Action Readout and Training}

DS-VLA stacks three such layers, with binary spikes passed between adjacent layers. Each layer contains 1536 neurons with two branches; the first receives the 768-dimensional modulated query. Continuous actions are decoded from the cumulative mean of the final-layer membrane state:
\begin{equation}
m_t=\frac{1}{t}\sum_{k=1}^{t}u_k^3,\quad
\widehat a_t=W_om_t+b_o.
\label{eq:action_readout}
\end{equation}
where $u_k^3$ is the final-layer membrane state, $m_t$ is its cumulative temporal mean, and $W_o,b_o$ are the learned continuous-action readout parameters. This readout retains subthreshold and temporal information unavailable in a purely binary spike representation. Neural states propagate across query positions, while their gradients are detached every eight internal steps to implement truncated backpropagation through time without resetting the states.

\begin{figure*}[t]
\centering
\includegraphics[width=\textwidth]{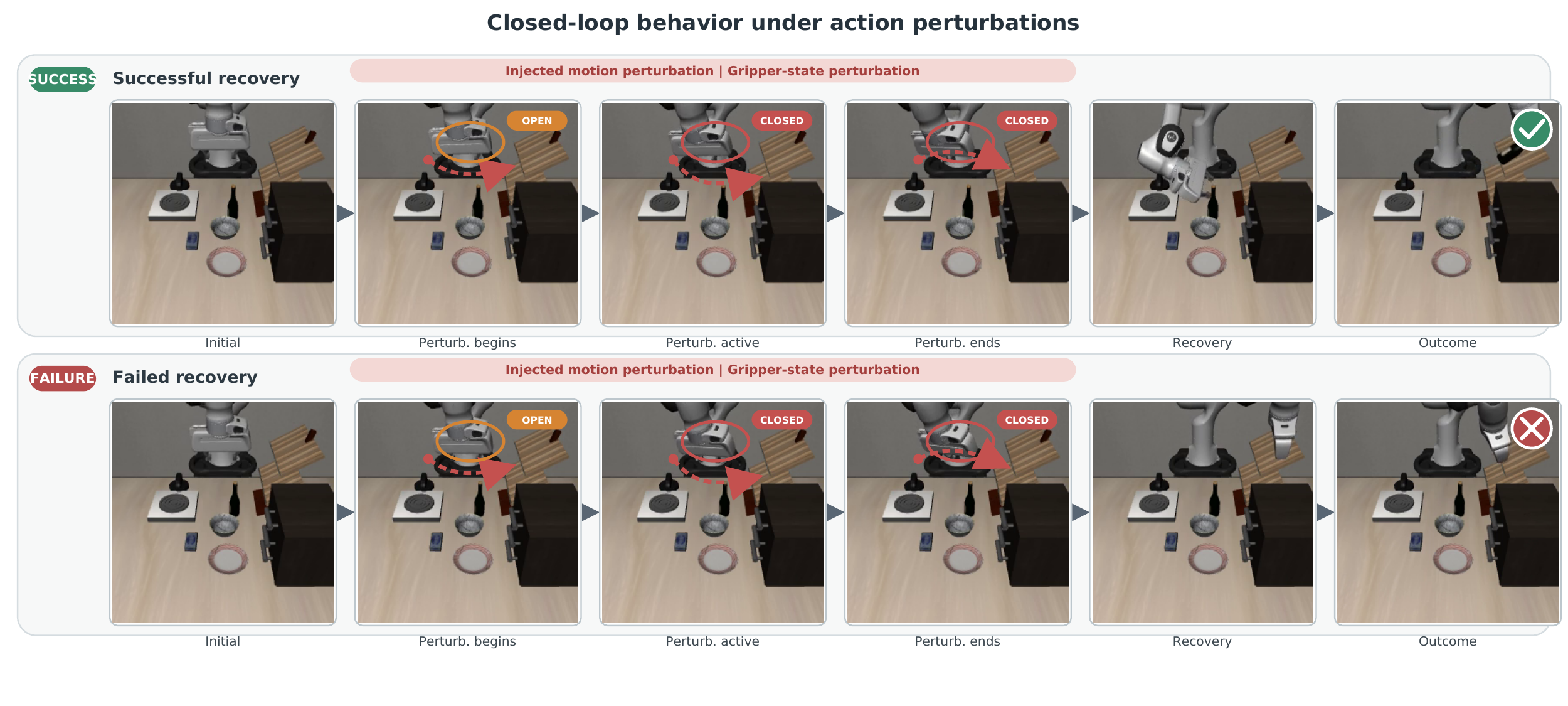}
\caption{Successful and failed DS-VLA trajectories under Difficulty~1. Red dashed arrows denote injected action displacement.}
\label{fig:perturbation_rollouts}
\end{figure*}

The model is optimized end to end with $L_1$ action regression, where $d_a=7$:
\begin{equation}
{\cal L}_{\rm action}
=\frac{1}{BTd_a}
\sum_{n=1}^{B}\sum_{t=1}^{T}
\left\|\widehat a_{n,t}-a_{n,t}\right\|_1.
\label{eq:action_loss}
\end{equation}
where $n$ indexes the $B$ batch elements, $t$ indexes the $T$ predicted actions, and $a_{n,t}$ is the ground-truth action. No auxiliary gate, firing-rate, or dendritic regularization loss is used.

\section{Experiments}
\label{sec:experiments}

We evaluate DS-VLA on standard manipulation, action perturbations of increasing severity, and inhibitory-gating ablations. Success rate is the primary metric, complemented by successful-trajectory and action-head analyses.

\subsection{Experimental Setup}

\subsubsection{Benchmark and Tasks}

We evaluate LIBERO-Spatial, LIBERO-Object, LIBERO-Goal, and Long, which emphasize spatial, object, goal, and long-horizon generalization~\cite{liu2023libero}. Each suite contains ten tasks with 50 episodes each, yielding 2,000 episodes.

Images are resized to $224\times224$, and the robot-state history contains 16 observations. Ten initial no-op steps stabilize the simulator and are excluded from the policy horizon. Maximum horizons are 220, 280, 300, and 520 for Spatial, Object, Goal, and Long; predicted actions are de-normalized to the Franka execution space.

\subsubsection{Baselines}

We compare with OpenVLA-OFT, FAST, $\pi_0$, and GR00T, competitive policies representing parallel continuous regression, autoregressive action tokens, flow matching, and dual-system diffusion control, respectively. Their distinct action dynamics complement the stateful dendritic and spiking dynamics of DS-VLA. All methods use the same task definitions, episode counts, success interface, and environment. The ungated variant isolates the contribution of inhibitory gating.

\subsubsection{Evaluation Metric}

Our primary metric is the task success rate
\begin{equation}
    \mathrm{SR}=\frac{N_{\mathrm{succ}}}{N_{\mathrm{all}}}\times 100\%,
\end{equation}
where $N_{\mathrm{succ}}$ and $N_{\mathrm{all}}$ denote successful and evaluated episodes. We aggregate tasks within each suite and macro-average the four suites.

\subsubsection{Action-Perturbation Protocol}

We perturb the executed 7-DoF actions. Difficulty~1 injects translation/rotation Gaussian noise ($0.1/0.002$) during steps 10--19 and flips the gripper command. Difficulties~2 and 3 extend the perturbation to 25 and 35 steps, increase the noise to $0.010/0.004$ and $0.015/0.008$, overwrite 10 and 12 translation commands with $[0.50,0,0]$ and $[0.65,-0.40,0]$, and force gripper closure for 18 and 25 steps, respectively. No perturbation indicator or hand-designed recovery signal is provided.

\subsection{Main Results}

DS-VLA achieves 91.6\% average success and 98.6\% on Object (Table~\ref{tab:clean_libero}). Although trailing OpenVLA-OFT, FAST, and $\pi_0$ on average, it remains competitive and outperforms GR00T, showing that dendritic spiking dynamics preserve strong nominal manipulation capability.

\begin{table}[!htbp]
\centering
\begin{tabular}{lccccc}
\toprule
Suite & DS-VLA & OFT & FAST & $\pi_0$ & GR00T \\
\midrule
Goal    & 94.6 & \textbf{97.9} & 96.1 & 95.8 & 86.0 \\
Object  & 98.6 & 98.4 & 97.2 & \textbf{98.8} & 92.0 \\
Spatial & 89.8 & \textbf{97.6} & 97.3 & 96.8 & 92.0 \\
Long    & 83.4 & \textbf{94.5} & 90.2 & 85.2 & 76.0 \\
\midrule
Average & 91.60 & \textbf{97.10} & 95.20 & 94.20 & 86.50 \\
\bottomrule
\end{tabular}
\caption{Success rate (\%) on standard LIBERO tasks. OFT denotes OpenVLA-OFT.}
\label{tab:clean_libero}
\end{table}

Standard success rates primarily evaluate nominal state--action trajectories and do not reveal recovery from execution-induced deviations. Clean performance alone is therefore insufficient to characterize closed-loop robustness.

Under Difficulty~1, DS-VLA achieves 87.35\% average success and retains 95.4\% of its clean performance (Table~\ref{tab:standard_perturb}). Its degradation is only 4.25 points, compared with at least 55.75 for the baselines, and it exceeds the strongest perturbed baseline by 47.9 points. On Long, DS-VLA reaches 77.8\% while all baselines remain at or below 1.2\%. Thus, DS-VLA combines competitive nominal execution with markedly stronger recovery from action-induced deviations.

\begin{table}[!htbp]
\centering
\begin{tabular}{lccccc}
\toprule
Suite & DS-VLA & OFT & FAST & $\pi_0$ & GR00T \\
\midrule
Goal    & \textbf{86.6} & 54.4 & 27.4 & 30.8 & 31.8 \\
Object  & \textbf{95.4} & 21.8 & 13.8 & 13.8 & 30.4 \\
Spatial & \textbf{89.6} & 80.4 & 54.4 & 68.6 & 60.8 \\
Long    & \textbf{77.8} & 1.2 & 0.0 & 1.0 & 0.0 \\
\midrule
Average & \textbf{87.35} & 39.45 & 23.90 & 28.55 & 30.75 \\
\bottomrule
\end{tabular}
\caption{Success rate (\%) under Difficulty~1 action perturbations. OFT denotes OpenVLA-OFT.}
\label{tab:standard_perturb}
\end{table}

Figure~\ref{fig:perturbation_rollouts} contrasts successful and failed responses on the same Goal task. The successful trajectory compensates after perturbation and completes placement; the failed trajectory does not recover before termination.

This robustness is consistent with the encoded biological principles. Dendritic compartments integrate signals locally with heterogeneous temporal responses, while dendrite-targeting inhibition strongly and selectively regulates branch excitability and somatic output~\cite{palmer2012inhibitory,goetz2015dendritic}. DS-VLA abstracts these mechanisms through branch-specific states and an input-dependent inhibitory coefficient shared by each neuron's two branches. The states preserve task-relevant information, while the gate limits abrupt updates without disrupting paired-branch organization. The clean and perturbed results therefore support combining dendritic temporal integration with neuron-wise inhibitory regulation.

\subsection{Robustness under Increasing Difficulty}

\begin{figure}[!htbp]
\centering
\includegraphics[width=\columnwidth]{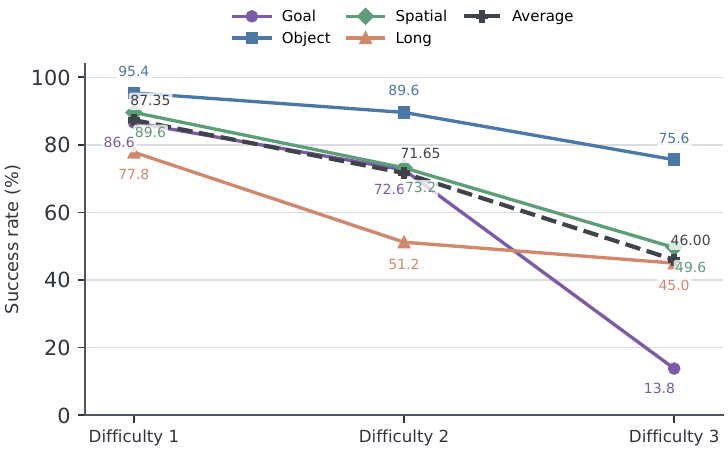}
\caption{DS-VLA success rates (\%) under increasing action-perturbation difficulty.}
\label{fig:difficulty}
\end{figure}

As disturbances become longer and more severe, DS-VLA's average success decreases from 87.35\% to 71.65\% and 46.00\% (Figure~\ref{fig:difficulty}), confirming the substantial challenge posed by these settings.

For successful trajectories, we analyze completion horizon, action frequency, inference and loop latency, and normalized first- and second-order action differences ($A$ and $J$), which are not metric Cartesian acceleration or jerk. As severity increases, the horizon grows from 145.4 to 175.4 steps, while action frequency, latency, and variation remain stable (Table~\ref{tab:success_difficulty}). Recovery therefore relies primarily on additional closed-loop corrections rather than increased per-step computation or action magnitude.

\begin{table}[!ht]
\centering
\setlength{\tabcolsep}{0.7mm}
\begin{tabular}{lrrrrrr}
\toprule
Level & Hor.$\downarrow$ & Hz$\uparrow$ & Inf.$\downarrow$ & Loop$\downarrow$ & $|A|\downarrow$ & $|J|\downarrow$ \\
\midrule
D1 & 145.4 & 6.56 & 122.6 & 151.2 & .029 & .040 \\
D2 & 155.2 & 6.52 & 123.8 & 152.7 & .027 & .037 \\
D3 & 175.4 & 6.73 & 119.9 & 147.8 & .028 & .039 \\
\bottomrule
\end{tabular}
\caption{Successful DS-VLA trajectories under increasing perturbation difficulty. Hor. is completion horizon; Inf. and Loop are in milliseconds.}
\label{tab:success_difficulty}
\end{table}

This profile is consistent with compartmentalized dendritic dynamics: branch-specific states and heterogeneous decay provide multiple timescales, while inhibitory gating limits abnormal overwriting. DS-VLA can thus preserve task context across repeated corrections without increasingly large or irregular actions. These success-conditioned statistics describe recovery rather than unconditional efficiency.

At Difficulty~3, forced actions may move the robot or object far from the nominal trajectory and create unrecoverable states. Dendritic dynamics and inhibitory gating correct moderate deviations but cannot resolve failures requiring a new task-level strategy. Stronger VLM failure recognition, recovery reasoning, and replanning should therefore complement the action head.

\subsection{Inhibitory-Gating Ablation}
\label{sec:gating_ablation}

\begin{figure}[!htbp]
\centering
\includegraphics[width=\columnwidth]{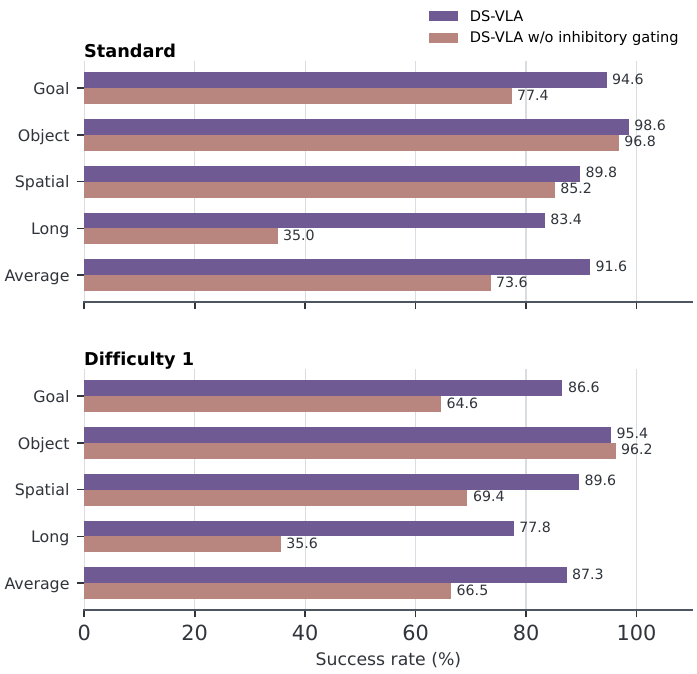}
\caption{Success rates (\%) for the inhibitory-gating ablation.}
\label{fig:gating_ablation}
\end{figure}

Removing inhibitory gating reduces clean and perturbed averages by 18.0 and 20.9 percentage points, respectively (Figure~\ref{fig:gating_ablation}). Because the ungated variant retains the same dendritic compartments and spiking dynamics, this gap isolates the additional value of input-dependent inhibitory regulation beyond dendritic temporal integration alone. The largest improvements occur on Long ($+48.4$ clean and $+42.2$ perturbed), indicating that selective state updating is particularly valuable when task information must persist over extended horizons. Perturbed Object is the sole exception, where the ungated variant is 0.8 percentage points higher.

\subsection{Action-Head Efficiency}

We compare the DS-VLA, OpenVLA-OFT, and GR00T action heads. FAST is excluded because it autoregressively generates action tokens within the VLM decoder and therefore has no separable action head for a like-for-like comparison. We report dense operations and an ideal event-driven estimate that skips fixed mask zeros and replaces spike-driven MACs with accumulations. Arithmetic energy is estimated by
\begin{equation}
E_{\rm arith}
=C_{\rm MAC}(\epsilon_{\rm mul}+\epsilon_{\rm add})
+C_{\rm ACC}\epsilon_{\rm add},
\label{eq:arithmetic_energy}
\end{equation}
where $C_{\rm MAC}$ and $C_{\rm ACC}$ count the corresponding operations. Following the CLIF implementation~\cite{huang2024clif} and Lemaire et al.'s framework~\cite{lemaire2022analytical}, we use $\epsilon_{\rm mul}=3.1$\,pJ and $\epsilon_{\rm add}=0.1$\,pJ. This operation-level proxy assumes firing rates of 0.10 and 0.05 and is not measured GPU energy.

\begin{table}[!ht]
\centering
\setlength{\tabcolsep}{0.7mm}
\begin{tabular}{lrrrr}
\toprule
Head & Params & Dense & Eff. MAC/ACC & Energy \\
\midrule
DS-VLA & 17.738M & 141.644M & 56.930M/3.052M & 0.182 \\
OFT & 42.004M & 335.774M & 335.774M/-- & 1.074 \\
GR00T & 100.865M & 78.327G & 78.327G/-- & 250.645 \\
\bottomrule
\end{tabular}
\caption{Analytical action-head cost. Energy is estimated arithmetic energy (mJ); effective operations assume ideal event-driven execution.}
\label{tab:action_head_cost}
\end{table}

DS-VLA uses 57.8\% fewer parameters and dense MACs than OFT. Under the ideal event-driven assumptions, its estimated arithmetic energy is 5.90$\times$ lower than OFT and 1,377.17$\times$ lower than ten-step GR00T, indicating substantial potential for sparse event-driven hardware.

On an RTX~4090 at batch size one, DS-VLA draws 104.27\,W on average, compared with 248.13\,W for OFT and 109.64\,W for GR00T. Spike-based communication and sparse dendritic connectivity may yield further event-driven hardware gains~\cite{davies2018loihi,blouw2019benchmarking}, depending on spike sparsity, hardware mapping, and native deployment.

\section{Conclusion}
\label{sec:conclusion}

We presented DS-VLA, a brain-inspired VLA model that translates dendritic computation and inhibitory regulation into robot control. DS-VLA partitions input across branches with distinct states and temporal responses, then integrates their activity through a spiking soma. Motivated by dendrite-targeting cortical inhibition, its neuron-wise shared gate regulates current entering two branches while preserving their states and organization. Rather than simulating complete biophysics, DS-VLA abstracts dendritic integration, temporal memory, somatic spiking, and inhibitory control. Together, these mechanisms determine how evidence enters the action state, persists, and how disruptive updates are constrained.

These biological priors benefit embodied control. DS-VLA achieves 91.6\% average success on standard LIBERO tasks and 87.35\% under Difficulty~1 action perturbations, retaining 95.4\% of its nominal performance and outperforming the strongest perturbed baseline by 47.9 percentage points. Removing inhibitory gating reduces clean and perturbed performance by 18.0 and 20.9 percentage points, respectively, showing that adaptive inhibitory regulation provides substantial gains beyond dendritic temporal integration alone. The spiking, sparse dendritic architecture also offers a path toward efficient event-driven execution. We therefore view DS-VLA as a step toward uniting brain-inspired computing and embodied intelligence: biological mechanisms can become actionable principles for building more robust, adaptive, and potentially energy-efficient robots. This perspective treats neuroscience not merely as inspiration, but as a source of testable mechanisms for robust and efficient embodied action generation under perturbations.

\bibliography{aaai2027}


\end{document}